\documentclass[11pt]{article}

\usepackage[preprint]{acl}

\usepackage{times}
\usepackage{latexsym}

\usepackage[T1]{fontenc}

\usepackage[utf8]{inputenc}

\usepackage{microtype}

\usepackage{inconsolata}

\usepackage{graphicx}

\title{EuroExec: Frontier Language Models Fall Short of Expert Judgment on European Executive Decision Tasks}

\author{
    \textbf{Pau Arnal}, \textbf{Khaled Denfir}, \textbf{Danylo Smahliuk},
    \textbf{Amrut Avhad}, \textbf{Marcus A. Castro} \\
    Sovrano AI \\
    \{pau, khaled, danylo, amrut, marcus\}@sovrano.ai
}

\begin{document}
\maketitle
\begin{abstract}

Frontier LLMs are increasingly put to use on open-ended complex questions, different in nature from the ones they are typically evaluated on. We dedicate more than 4,000 human expert hours to evaluate a selection of six frontier LLMs on a member of this class of problems: EuroExec, our introduced human expert-based benchmark composed of 413 open-ended long-form European executive tasks authored by 47 vetted domain experts, each question drawn from experience in a real case. Every response is manually evaluated through a multi-attribute rubric, an item-specific checklist of requirements, and a preference rank ordering, extracting an aggregate metric ``Solve Rate''. The strongest model solves only 56.9\% of tasks, while expert-written reference answers judged blindly are solved at near-ceiling levels and are preferred over every model response in 74\% of direct rankings, placing frontier generative systems well below the professional standard of work they are already used for. We see that the best way to extract this kind of conclusion is by employing human evaluators, carefully checking their consistency through rigorous statistical analysis, and observe that automatic measurements also fall short when evaluating on this case of real-world open-ended problems with a subjective ground truth.

\end{abstract}  

\section{Introduction}

Evaluation of automatic text generation systems is fundamental to the advancement of the field of generative models, key for both their development and benchmarking \citep{reiter-2026-nlg}. This has typically been sought by carefully designing tasks for objective deterministic metrics to be applicable, such as multiple-choice problems \citep{zellers-etal-2019-hellaswag, hendrycks2021measuring, rein2024gpqa}, tasks where the output is restricted in other ways \citep{rajpurkar-etal-2016-squad, kwiatkowski-etal-2019-natural}, tasks with gold-standard responses such as summarization or translation \citep{papineni-etal-2002-bleu, lin-2004-rouge, Zhang2020BERTScore}, or programming exercises where the output can be executed and automatically evaluated \citep{chen2021evaluatinglargelanguagemodels, ICLR2024_edac78c3, ICLR2025_94074dd5}. Collections of such close-ended generation tasks have also been employed to measure the general capabilities of generative models \citep{NEURIPS2019_4496bf24, wang2018glue, srivastava2023bigbench, liang2023holistic, eleuther-eval-harness}.

These approaches serve to evaluate a model's inherent knowledge, reasoning, or multimodal capabilities, but do not directly measure its performance on the kinds of open-ended text generation tasks LLMs are routinely put to work on \citep{novikova-etal-2017-need, huang2023ceval}, like chat interactions, general problem solving, creative drafting, documentation analysis, or as in the case of this work executive decision-making assistance. In all of the aforementioned applications and others like them, there is no single perfect answer. Nonetheless, it does have a professional standard, and practitioners in the field can recognize when it is met \citep{basile-etal-2021-need, uma2022disagreement}. Not only are open-ended tasks more representative of the real uses of generative models, but the lack of a ``golden answer'' has other benefits: the field has needed to grapple with the fact that all of the public benchmarking tasks with definite correct answers may have been trained on by the models they are supposed to evaluate, leaving the door open to question how much the improvement shown by newer models is due to better base performance and how much it is due to their regurgitation of answers they already know \citep{zhou2023dontmakellmevaluation, yang2023rethinkingbenchmarkcontaminationlanguage, Li_Flanigan_2024, zhao-etal-2025-mmlu-contamination}.

Recently, some approaches to evaluation of open-ended text generation tasks leveraged LLMs themselves as AI judges following a preset rubric \citep{chiang-lee-2023-large, chen-etal-2023-exploring-use, desmond2024evalullm, bavaresco-etal-2025-llms, wang2025canreplace, akyurek-etal-2026-prbench}, but the question remains open as to how much these LLM judges are truly representative of the standard they are purported to represent: actual human judgment, especially expert-based evaluation \citep{zheng2023judgingllm, van-miltenburg-etal-2023-reproducible, chen-etal-2024-humans, NEURIPS2024_7f1f0218, szimanski2025limitations, ICLR2025_9e720fce, huidrom-belz-2025-ask, ni-etal-2026-reasoning}. Due to the cost involved, only a few recent studies have involved a full human-based evaluation of an open-ended text generation task \citep{chiang2024chatbotarena, guo2026superconductivity, badawi2026assessingqualitymentalhealth}.

We introduce EuroExec, a human-based benchmark focused on expert evaluation of open-ended long-form questions for European executive-grade use cases, with a collection of 413 detailed and complex open-ended questions authored by experts with certified professional experience, drawing on cases they have worked on. The answers generated by different models were evaluated manually by these experts using three instruments: a common rubric with several categories, a specific checklist of requirements authored jointly with the question, and an explicit preference ranking of all responses. We observe that all frontier models fall short of the standards required to assist European executive decision-making, standards which the experts fulfill by a comfortable margin, and that the best way to assess frontier models on these kinds of tasks is founded in human evaluation.

Throughout the rest of the manuscript, we first detail how the question dataset was built (Section~\ref{sec:question-dataset}), how we gathered the automatic responses (Section~\ref{sec:response-generation}), and the methodology followed for the expert evaluation (Section~\ref{sec:expert-evaluation}). We follow with a discussion of the results in Section~\ref{sec:results} and conclude in Section~\ref{sec:conclusion}.

\section{Question Dataset}
\label{sec:question-dataset}

We selected 47 experts through CV screening and one-on-one interviews. The experts were divided into four domains based on their expertise: Finance, Marketing, Business, and Product. All of them have business and management backgrounds specific for Europe with first-hand professional experience in the domain in which they author questions and evaluate model responses. Table~\ref{tab:categories} contains a summary of the number of experts and questions for each domain.

\subsection{Question Formulation}

Each question item in the dataset was manually crafted as a question description with a checklist that the answer should cover. Question authors were given instructions so that the question must be: self-contained (pure reasoning task), open-ended, long-form, specific (specific place(s), plausible names, at least one friction point), goal-oriented (state an objective without giving any steps, the difficulty comes from the judgment), and European market-specific by design (for example a works-council step in a German HR decision). The question authors were given freedom to choose topics and specific situations they know first-hand from their own professional experience, where they are able to determine with certainty the essential aspects any correct answer must fulfill. The questions describe real business situations in considerable detail, and tend to be long ($\sim$1,200 characters, $\sim$200 words) and require an expansive response. The checklist, created together with the question, plays the role of a ground truth: each checklist contains 5 to 10 verb-first phrases describing criteria that a good response would fulfill. See Appendix~\ref{app:example-item} for a graded example of such a question item.

\begin{table}
    \centering
    \begin{tabular}{lcc}
    \hline
    \textbf{Category} & \textbf{Experts} & \textbf{Questions} \\
    \hline
    Finance & 10 & 74 \\
    Marketing & 11 & 85 \\
    Business & 14 & 113 \\
    Product & 12 & 141 \\
    \hline
    \textbf{Total} & \textbf{47} & \textbf{413} \\
    \hline
    \end{tabular}
    \caption{Number of experts and questions for each domain}
    \label{tab:categories}
\end{table}

\subsection{Validation Pipeline}
\label{ssec:valid-pipeline}

To ensure the relevance and quality of the questions so that they are suitable for evaluating frontier generative models, they were passed through a specifically designed validation pipeline.

The first stage or ``Automated QA'' consisted of an automatic LLM (Claude Sonnet 4.6) checking the question-checklist item pair on 6 rejection criteria: hint leakage (the question reveals the checklist items), checklist relevance (the checklist matches the question), unwanted patterns (identity probing, injection, unsafe prompt or question not self-contained), unprofessional language, unclear role, or non-specificity of the scenario.

The second stage or ``Difficulty Gate'' involved two LLMs from different model families (to avoid the self-enhancement bias found by \citet{NEURIPS2024_7f1f0218}), one of them acting as {\it solver} (Claude Haiku 4.5) and another as {\it grader} (Gemini Flash 3.5), to ensure all questions in the dataset present some reasonable judgment or reasoning challenge to current frontier models. The {\it grader} cross-referenced the automatic answer of the model against the checklist, and the difficulty score is defined by the percentage of automatically passed checklist facts. A maximum of 60\% first-pass automatic fulfillment rate was established from an average of 5 passes as a threshold for the questions. While this step did involve automatic grading of questions as part of the process, it was needed for the question authors to rapidly iterate, and does not detract from the reliability of human over automatic evaluations as we will see in Section~\ref{ssec:automatic-metrics}. The authors were asked to rewrite either the question, the checklist, or both, until an automatic LLM response would not automatically fulfill the checklist, which took on average $\sim$30 attempts. We used the scores from the ``Difficulty Gate'' to categorize the final question items into Easy, Medium, and Hard, without targeting any specific distribution. Table \ref{tab:difficulties} shows the number of question items in our handcrafted dataset for each difficulty according to their automatic checklist fail rate. See Appendix~\ref{app:validation-steps} for examples of questions automatically rejected by this validation pipeline.

\begin{table}[]  
    \centering
    \begin{tabular}{lcc}
        \hline
        \textbf{Difficulty} & \textbf{Checkl. (\%)} & \textbf{Questions} \\
        \hline
        Hard      & $[0, 30]$  & 100 \\
        Medium    & $(30, 50]$  & 137 \\
        Easy      & $(50, 60]$  & 176 \\
        \hline
    \end{tabular}
    \caption{Number of questions for each difficulty category resulting from the ``Difficulty Gate'' scores}
    \label{tab:difficulties}
\end{table}

\section{Response Generation}
\label{sec:response-generation}

We selected six frontier LLMs, chosen to span the leading commercial systems and cover multiple providers from all parts of the world, which are listed in Table~\ref{tab:models}. Each model was queried through its provider API using its own default decoding settings. The models were passed the questions as a single prompt without any additional context such as the system prompt, conversation history, tools, or file access, consistent with the self-contained design of the items. Cost and response length were recorded per response for a secondary analysis contained in Appendices~\ref{app:cost} and \ref{app:length}.

\begin{table}[h]
    \centering
    \begin{tabular}{ll}
    \hline
    \textbf{Model} & \textbf{Provider} \\
    \hline
    Fable 5 & Anthropic \\
    Claude Opus 4.8 & Anthropic \\
    GPT-5.5 & OpenAI \\
    Gemini 3.1 Pro & Google \\
    GLM-5.2 & Zhipu AI \\
    Mistral Large & Mistral AI \\
    \hline
    \end{tabular}
    \caption{The six evaluated frontier models and their providers.}
    \label{tab:models}
\end{table}

\section{Expert Evaluation}
\label{sec:expert-evaluation}

The evaluations were performed according to three independent instruments: scores for a common rubric, item-specific checklist fulfillment, and an explicit ranking of all responses according to preference. The checklist criteria were manually evaluated as hit, partial, or miss. The rubric consisted of the same five attributes for all questions, scored 1 to 5 on a Likert scale:

\begin{table*}[t]  
    \centering
    \begin{tabular}{l|ccccc}
        \hline
        \textbf{Model} & \textbf{Solve Rate} & \textbf{Avg. Rank} & \textbf{Win (\%)} & \textbf{Rubric} & \textbf{Checkl. (\%)}  \\
        \hline
        Human Expert*   & \textbf{92.4} & 1.76 & 74.24 & 4.33 & 94.9 \\
        \hline
        Fable 5         & \textbf{56.9} & 2.13 & 49.5 & 4.14 & 61.9 \\
        GPT-5.5         & 51.3 & 2.85 & 22.5 & 3.91 & 58.4 \\
        Opus 4.8        & 34.1 & 3.24 & 9.4 & 3.78 & 51.5 \\
        Gemini 3.1 Pro  & 26.9 & 3.73 & 7.9 & 3.64 & 47.5 \\
        GLM-5.2         & 21.1 & 4.18 & 5.0 & 3.41 & 42.9 \\
        Mistral Large   & 18.4 & 4.87 & 5.7 & 3.04 & 37.8 \\
        \hline
    \end{tabular}
    \caption{Summary of results from different human evaluation methods: Solve Rate, preference ordering (average rank and win rate), rubric score, and checklist fulfillment rate. Human results from a separate 33-question subsample.}
    \label{tab:headline}
\end{table*}

\paragraph{Domain} Is the information presented in the answer adequate to the domain of the question? Are the facts, figures, and frameworks presented correct?

\paragraph{Localization} Does the response follow market-specific local customs, rules, institutions, and regulations?

\paragraph{Reasoning} Is the logic sound and leads to the presented recommendation, which respects the constraints specified by the question?

\paragraph{Communication} Is the response oriented towards a recommendation, clearly structured, and concise for an executive audience?

\paragraph{Actionability} Does the answer present concrete and realistic plans, accounting for the sequence of actions, and including specific values, dates, and tools?

\paragraph{}
During human evaluation, model responses were anonymized and shuffled for each question. Each response was independently graded by two domain-specific evaluators. This is a highly time-intensive process: with responses averaging around 44,000 characters ($\sim$7,500 words), their evaluation along the three instruments takes around 5 hours per item for all six responses. More than 4,000 human hours were dedicated to evaluating on the full 413 questions. In addition, for a subset of 33 questions, the question authors were asked to write out an ideal response, which was evaluated blindly as a seventh model by two independent experts.

From all three instruments, we extract an overview metric representing the average performance of every model: we introduce the ``Solve Rate'' (SR) aggregate metric, defined by the share of question items it \textit{solves}. A question item is considered \textit{solved} when the mean rubric score is $\geq 3.0$ with a checklist fulfillment rate $\geq 60\%$, which we consider to be a generous passing grade for a supposedly strict examination. Other metrics extracted were: average rank from preference ordering, win rate (the percentage of times the model produced the most preferred response), mean rubric score, and checklist fulfillment rate (for which we take a \textit{partial} assessment as 0.5).

\section{Results}
\label{sec:results}


We start by presenting overall scores and later discuss individual instruments separately and how we used them together to validate intra- and inter-evaluator consistency. We finish by comparing our expert-based evaluation methodology with the most common types of automatic metric.

Starting with the overall scores, Table~\ref{tab:headline} shows five different metrics extracted from the three evaluation instruments. All five measures mostly agree with the same ordering, a strong indicator towards consistency. We can see a stronger dropoff in win rate (the fraction of question items to which the model generated the first-ranked response) than in the rest of the metrics; this is to be expected as it is the most competitive metric where only one model can ``win'' a question, whereas we consider other metrics such as Solve Rate more indicative of the state of each model. An extra row at the top was extracted from an equivalent table for the 33-question subset containing human expert-written answers, found in Appendix~\ref{app:subsample-expert-answers}. We can immediately see the impressive gap that qualified human experts still have with frontier generative models, especially in the Solve Rate and checklist fulfillment rates.

\begin{figure}
    \centering
    \includegraphics[width=\linewidth]{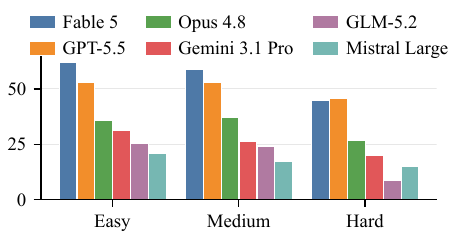}
    \caption{Solve Rate (\%) broken down by question difficulty from the Difficulty Gate}
    \label{fig:solve-rate-by-difficulty}
\end{figure}

\begin{figure*}  
    \centering
    \includegraphics[width=\linewidth]{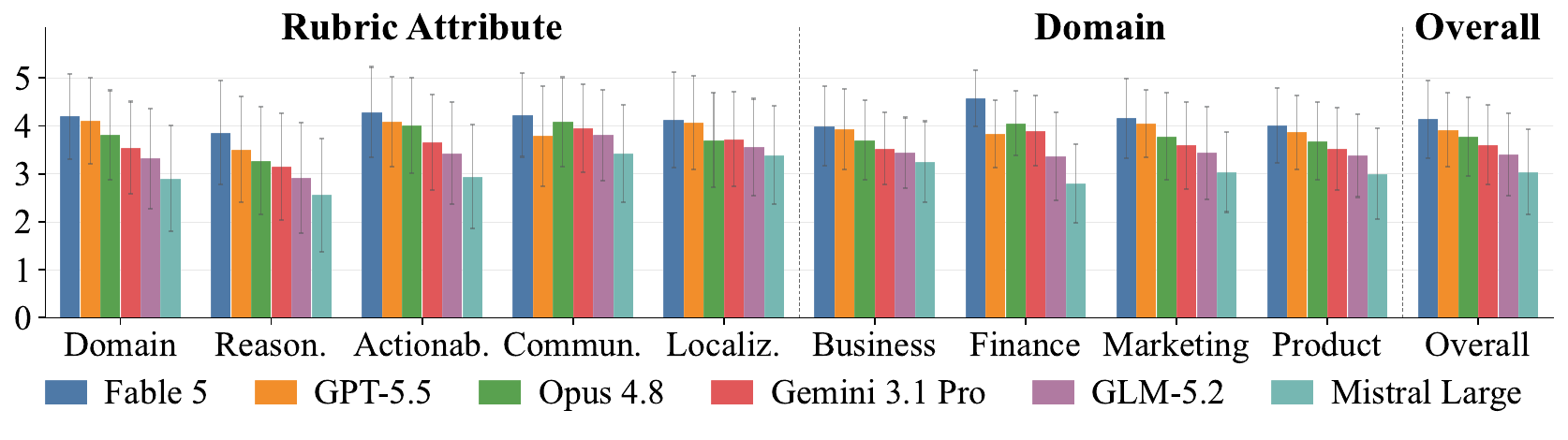}
    \caption{Rubric scores (mean $\pm$ std dev), computed by attribute, aggregated by domain, and overall scores}
    \label{fig:rubric-scores}
\end{figure*}

Returning to the comparison among models, Figure~\ref{fig:solve-rate-by-difficulty} contains a breakdown of Solve Rate by difficulty of the questions according to the automatic ``difficulty gate'' (see Section~\ref{ssec:valid-pipeline}). We can see a general trend of stronger models performing better across all difficulties, which breaks down somewhat in the hardest difficulty, notably with the best overall model (Fable 5) having a larger relative degradation and ending up in the second position of the ranking. A noticeable observation is that even the most capable frontier models only barely surpass the 50\% Solve Rate in the easiest subset, coming again to the conclusion that all of them are far from the standards needed to assist in European executive decision-making.


\subsection{Rubric-based Evaluation}
\label{ssec:rubric-based-evaluation}

Figure~\ref{fig:rubric-scores} reports mean and standard deviation for each of the five rubric attributes, computed over all items. We also aggregated the mean rubric scores for each domain, and computed the overall score. We can observe at a glance that all of the evaluated dimensions are strongly linked to each other in these ---admittedly generalist--- models. 

Looking at rubric attribute scores, we see stronger floors in Localization and in Communication, showing that LLMs generally shine more at text structure and communication, while the worst performance across the board is seen in Reasoning. In a way, this reflects that even today's frontier LLMs are still measurably better at generating fanciful text than true intelligent reasoning. GPT-5.5 in particular obtained relatively poor assessments in Communication, see Appendix~\ref{app:length} for our interpretation. The next section of the figure breaks down the rubric scores along the four domains of our evaluation benchmark: Business, Finance, Marketing, and Product. Once again, we can observe a strong relationship in performance across domains, with a singular exception in Finance, where GPT-5.5 is not at the same relative level as displayed in the other domains. This anomalous result can represent a relatively poorer numerical capability, which more strongly affects financial problems.

\begin{figure}
    \centering
    \includegraphics[width=\linewidth]{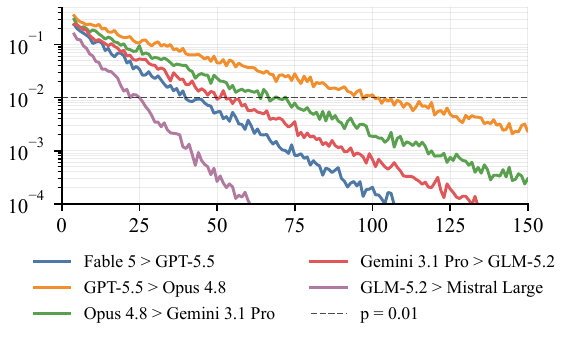}
    \caption{Median p-value (500 resamples) from paired t-test on mean rubric scores as a function of sample size}
    \label{fig:p-values}
\end{figure}

Although the model ranking derived from the rubric seems mostly consistent across rubric attributes and domains, all standard deviations are relatively high. To verify that the overall ranking from the rubric score is significant, we performed standard paired t-tests from each model to the next in the ranking, obtaining p-values ranging between 1.26e-6 and 9.2e-23, representing extremely high statistical significances for all pairwise orderings. An interesting question that naturally arises from this analysis is: what would the minimum number of questions needed to reliably establish such a ranking have been? Figure \ref{fig:p-values} plots the median (out of 500 random resamples) p-value of paired t-test as a function of sample size. We can see that all pairs cross $p=0.01$ at 100 samples, and 150 questions is more than enough to establish the ranking at a high level of statistical significance. Outside of the plot, the lines follow the trend, crossing $p=0.001$ at $n=179$ and going on until obtaining the final 1.26e-6 at $n=413$, extremes which were left out of the plot for readability reasons. For the 33-question subset including human answers, experts were preferred over any LLM with $p<0.01$, a reassuring significance level for the small sample.

\begin{table}  
    \centering
    \begin{tabular}{l|c}
        \hline
        \textbf{Attribute} & \textbf{$r$ w.r. to others} \\
        \hline
        Domain         & 0.81 \\
        Reasoning      & 0.68 \\
        Actionability  & 0.82 \\
        Communication  & 0.63 \\
        Localization   & 0.75 \\
        \hline
        Checklist      & 0.69 \\
        \hline
    \end{tabular}
    \caption{Pearson's correlation coefficient $r$ of each attribute's score with respect to the mean of the rest. An additional row shows the correlation of the checklist fulfillment rate to the overall rubric score}
    \label{tab:rubric-correlation}
\end{table}

A common theme of these results is that an improvement in one aspect of a model's performance seems to be accompanied by an overall improvement across the board. To further probe into this observation, we computed the Pearson's correlation coefficient between every attribute and the overall score of the rest of attributes from every individual grading, as presented in Table \ref{tab:rubric-correlation}, where we also included the checklist fulfillment rate against the overall rubric score. We can see that they are all strongly positively correlated to each other, while retaining some independent signals, the strongest of them being Reasoning and Communication both with $r<0.7$. Meanwhile, Domain and Actionability were the most generic-aligned attributes with $r\geq0.8$. The most distant to the overall scores was Communication, attribute most related to the perception of the quality and succinctness of the summary, perhaps because a shorter answer may be positively evaluated in this dimension while penalized with respect to the rest and vice versa.

\begin{figure}  
    \centering
    \includegraphics[width=\linewidth]{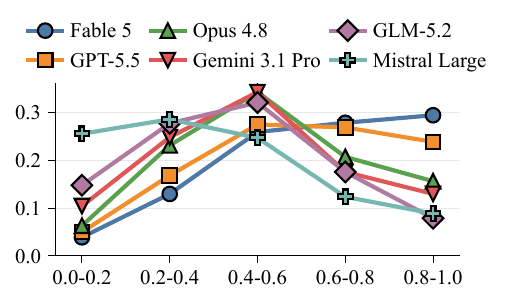}
    \caption{Proportion of checklist evaluations within each range of fulfillment rates, for each evaluated model, in ranges of size 0.2}
    \label{fig:checklist-cov-dist}
\end{figure}

\subsection{Checklist-based Evaluation}

While the rubric score is more representative of a subjective perception, we also asked the experts to analyze objectively if the answer fulfills the checklist requirements as established by the author of the question, obtaining checklist fulfillment rate figures for every question, which are indicative of how close each LLM has been to an objectively correct answer as stated by the author.

In particular, we are interested in the distribution of these checklist fulfillment rates: from total failures (0.0) to perfect answers (1.0), how are they distributed among the responses of each model? Figure \ref{fig:checklist-cov-dist} depicts this distribution across five ranges of size 0.2. In general, we can see quite a wide distribution over all the question items, showing that they cover a wide range of difficulties. Looking at specific lines, we can see weaker models such as Mistral Large with most of their distribution skewed to the lower-end values, while stronger models such as Fable 5 or GPT-5.5 have more representation in the highest fulfillment rates.

Checklist fulfillment rates are used to compute the Solve Rate aggregate metric and play a major role in the analysis of evaluator consistency that follows in the next section.

\subsection{Evaluator Consistency}

\begin{figure*}[t]
    \centering
    \includegraphics[width=\linewidth]{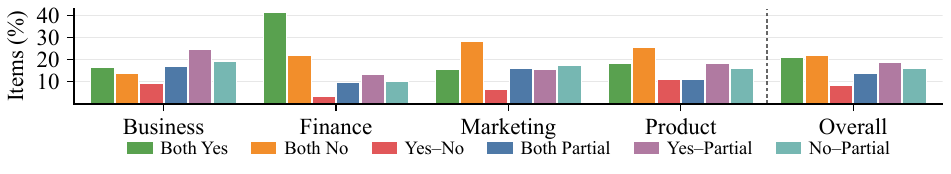}
    \caption{Breakdown of annotator pair assessments at the level of individual checklist item, by domain and overall}
    \label{fig:checklist-agreement}
\end{figure*}

In subsection~\ref{ssec:rubric-based-evaluation}, we have already partially discussed the correlation between the checklist fulfillment rate and the rubric scores as a measure of internal coherence of the evaluations when analyzing Table~\ref{tab:rubric-correlation}, to conclude that they were consistent signals with independent components. In addition to the rubric and checklist evaluations, the participants were asked to rank the generated answers according to their preference. We leverage this data to extract correlations of the explicit rankings with others derived from the rubric scores and the checklist fulfillment rate. To estimate the correlations between ordered rankings, we used Kendall's coefficient $\tau$.

\begin{table}[]  
    \centering
    \begin{tabular}{lc}
        \hline
        \textbf{Comparison} & \textbf{Kendall $\tau$} \\
        \hline
        explicit vs rubric       & 0.77 \\
        explicit vs checklist    & 0.67 \\
        checklist vs rubric      & 0.68 \\
        \hline
        cross-eval. explicit   & 0.34 \\
        cross-eval. rubric     & 0.35 \\
        cross-eval. checklist  & 0.43 \\
        \hline
    \end{tabular}
    \caption{Kendall rank correlation coefficient $\tau$ computed over explicit and derived rankings, internally for each evaluator (top) and across evaluators (second)}
    \label{tab:kendall-tau}
\end{table}

Table~\ref{tab:kendall-tau} contains the values of the $\tau$ coefficients for all internal consistency comparisons between the three instruments, and other three cross-evaluator comparisons again on all of them. All values are significantly positive, which indicates a clear statistical relationship between any pair of rankings, explicit or derived. We also see a clear difference between internal consistency and cross-evaluator consistency, with the former being higher as expected. An interesting observation is that subjective evaluations such as preference ordering and rubric-based ordering correlate better with each other than either does to the more objective-oriented evaluation of the checklist, while across evaluators the highest correlation is also observed on the checklist, corroborating that the results from the checklist are somewhat more independent of the particular evaluator than the purely subjective metrics. This is again clearly observed when extracting the Pearson correlation coefficient $r$ between all pairs of evaluations on rubric score ($r=0.29$) and checklist fulfillment rate ($r=0.52$).

Relatedly, another indicator of inter-annotator agreement can be seen at the level of individual checklist item. Figure~\ref{fig:checklist-agreement} contains a breakdown of every possible pair of assessments for each checklist item, divided by domain and overall. The lowest number in all cases is the ``yes-no'' combination, and interestingly this is sensitive to the domain, with Product having the highest ``yes-no'' rate and Finance the lowest. This may reflect this latter domain being just easier overall, looking at the ``both yes'' rate; the same can be said for Marketing where there is also a low ``yes-no'' combination but in this case the domain is harder overall leading to a relatively low disagreement rate. In general, the figure shows that checklist items tend to have a low disagreement rate when checking the ``yes-no'' combination, retaining significant nuance involving the ``partial'' assessment. 

\begin{figure}[]
    \centering
    \includegraphics[width=\linewidth]{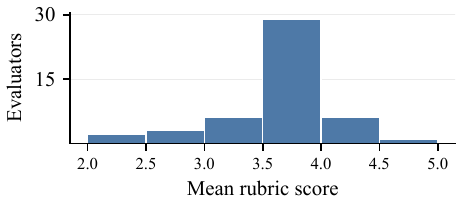}
    \caption{Number of evaluators according to their own mean rubric scores to depict the leniency distribution}
    \label{fig:histogram-evaluators}
\end{figure}

\begin{figure}[]
    \centering
    \includegraphics[width=\linewidth]{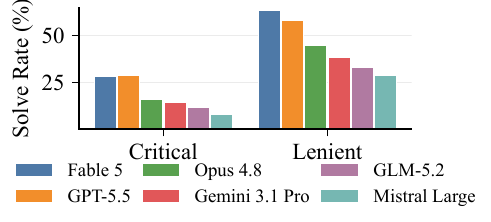}
    \caption{Solve Rate according to the 10 most critical and 10 most lenient annotators}
    \label{fig:solve-rate-leniency}
\end{figure}

Inter-evaluator consistency can also be gauged as a group. Figure~\ref{fig:histogram-evaluators} contains a histogram of the number of evaluators according to the mean rubric scores from their gradings. We can see that the resulting distribution is more concentrated than a normal distribution, with the great majority having a mean rubric score between 3.5 and 4.0 points. We computed Solve Rates according to the 10 most critical and the 10 most lenient evaluators, as shown in Figure~\ref{fig:solve-rate-leniency}, and obtain rankings fully consistent with the rest of the paper, even if Fable 5 and GPT-5.5 are confounded for the critical group, this may well be noise due to the small sample size. This shows that even the ``unreasonably'' harsh or lenient evaluators are contributing valuable information and remain coherent with the rest. A secondary observation is that once again, even taking only the most lenient evaluators, the Solve Rate that we can derive from them for the most capable frontier models only barely surpasses 50\%.

\subsection{Expert Evaluation vs Automatic Metrics}
\label{ssec:automatic-metrics}

We compared our human-based evaluation methodology against the main automatic approaches that have been used for similar tasks, namely deterministic scores using a reference and LLM judges.

For the former, we computed ROUGE-Lsum and BLEU for a subset of 32 questions with expert-written answers equally distributed among the 4 domains, as shown in Table~\ref{tab:autometrics} alongside their average human rubric score. Some positive correlation does exist between these automatic metrics and the mean human rubric score at the item level (BLEU having on average Pearson $r=0.27$ and Kendall $\tau=0.21$ with the mean rubric score, and ROUGE-Lsum having $r=0.37$ and $\tau=0.29$), perhaps due to the usage of certain ``right'' words or short phrases, but it does not translate to the ranking. This shows that these kinds of automatic metrics are not useful for determining the capabilities of frontier models on real-world executive decision-making.

\begin{table}[]
    \centering
    \begin{tabular}{l|ccc}
        \hline
        \textbf{Model} & \textbf{Rubric} & \textbf{ROUGE} & \textbf{BLEU} \\
        \hline
        Fable 5         & 4.15 & 0.43 & 3.7 \\
        GPT-5.5         & 3.78 & 0.41 & 3.5 \\
        Opus 4.8        & 3.68 & 0.43 & 4.1 \\
        Gemini 3.1 Pro  & 3.57 & 0.41 & 3.5 \\
        GLM-5.2         & 3.30 & 0.41 & 3.5 \\
        Mistral Large   & 3.10 & 0.36 & 2.4 \\
        \hline
    \end{tabular}
    \caption{Mean rubric, ROUGE-Lsum (F1), and BLEU scores for 32 balanced questions with golden answers}
    \label{tab:autometrics}
\end{table}

For the case of LLM-based AI judges, we used DeepSeek v4-Pro as an independent judge \citep{NEURIPS2024_7f1f0218} by asking it to fill in the three evaluation instruments for the full dataset, and we get a strong correlation with human scores. As seen in Table~\ref{tab:kendall-tau-llm} and cross-checking with Table~\ref{tab:kendall-tau}, we are able to replicate the results of \citet{huidrom-belz-2025-ask} where LLM-based judges correlate better with human evaluators than human evaluators do with each other. Superficially, this can lead to the wrong-headed conclusion that AI judges are more representative of the average human than a single evaluator. When looking at the actual results shown in Table~\ref{tab:llm-judge-scores}, we see that the LLM over-estimates the best performing models and under-estimates the worst, outputting a more variable score that obtains a stronger correlation signal as an artifact. Not only do they fail to capture the diversity of human opinions which contains valuable information as demonstrated by \citet{ni-etal-2026-reasoning}, but they also critically fail to accurately replicate the subtlety of human evaluation, outputting too crisp evaluations.

\begin{table}[]  
    \centering
    \begin{tabular}{lc}
        \hline
        \textbf{Comparison} & \textbf{Kendall $\tau$} \\
        \hline
        LLM-indiv. explicit   & 0.47 \\
        LLM-indiv. rubric     & 0.48 \\
        LLM-indiv. checklist  & 0.51 \\
        \hline
        LLM-mean explicit   & 0.54 \\
        LLM-mean rubric     & 0.54 \\
        LLM-mean checklist  & 0.55 \\
        \hline
    \end{tabular}
    \caption{Correlations with orderings derived from automatic LLM judges. Above, Kendall $\tau$ between LLM and individual humans. Below, between LLM and the mean of the humans}
    \label{tab:kendall-tau-llm}
\end{table}

\begin{table}[t]  
    \small
    \centering
    \begin{tabular}{l|cc|cc}
        \hline
                        & \multicolumn{2}{c|}{\textbf{Human}} & \multicolumn{2}{c}{\textbf{LLM}} \\
        \textbf{Model}  & \textbf{Rubric} & \textbf{Chkl.} & \textbf{Rubric} & \textbf{Chkl.}  \\
        \hline
        Fable 5         & 4.14 & 61.9 & 4.54 & 61.0 \\
        GPT-5.5         & 3.91 & 58.4 & 4.11 & 51.1 \\
        Opus 4.8        & 3.78 & 51.5 & 3.86 & 47.1 \\
        Gemini 3.1 Pro  & 3.64 & 47.5 & 3.45 & 39.9 \\
        GLM-5.2         & 3.41 & 42.9 & 3.22 & 34.5 \\
        Mistral Large   & 3.04 & 37.8 & 2.72 & 27.2 \\
        \hline
    \end{tabular}
    \caption{Rubric score and checklist fulfillment rate as assessed by human evaluators and LLM judge}
    \label{tab:llm-judge-scores}
\end{table}

\section{Conclusion}
\label{sec:conclusion}

We evaluate the performance of a selection of 6 frontier models on real-world European executive decision-making as a member of a class of long-form open-ended problems which LLMs are routinely put to work on and marketed towards. More than 4,000 hours of human experts have been dedicated to evaluating the models on 413 complex questions belonging to four domains using three independent evaluation instruments. We draw two main conclusions from this work: that even the top-performing frontier models as of today fall short of the standards required for this real-world open-ended task, and that in these kinds of tasks human evaluation is still unmatched by any other kind of automatic measurement.

We see that even setting generous conditions for a question to be considered ``solved'' by a response (60\% checklist fulfillment rate, 3/5 in rubric score), even the top-performing models just barely scratch the 50\% Solve Rate while human experts ace the test comfortably over the 90\% Solve Rate. Human evaluations are diverse but consistent on issues without a single objective answer but with real professional standards to comply with, producing almost identical rankings from a wide variety of decompositions, obtaining a reassuring worst-case \mbox{p-value} for the rankings of 1.26e-6.

In comparison with automatic metrics, we observe that human assessment is unmatched by either old-school deterministic scores comparing against a golden answer or flashier LLMs acting as AI judges, which apart from the self-inflating bias described by \citet{NEURIPS2024_7f1f0218}, also fail to replicate the nuances and heterogeneity of human evaluations, missing critical information and not being truly representative of the average human, against some claims in the literature.

\section*{Limitations}
Although we show human evaluation is the best way to assess generative models on open-ended tasks, this approach has inherent limitations as any human effort has to be planned and coordinated, and must be carefully prioritized. As a consequence, we only had the expert-generated ideal answer for 33 of the 413 questions, muddling the interpretability of some results to a degree. Another clear limitation is found in the reproducibility of the evaluation, notwithstanding the cost: not only would another batch of evaluators generate different numbers that would need to be statistically cross-checked, but as time passes extending the benchmark to newer models becomes incrementally infeasible without replicating the full study as the organization loses access to the time of the specific experts involved.

We left a qualitative analysis of failure modes outside the scope of this work in favor of a thorough quantitative analysis of the reliability of human evaluation and its comparison to automatic metrics, but it is a dimension that should be included in future work, and it could become an especially interesting contribution if we can design approaches to employ the huge amounts of coordinated human effort described in this work for such a purpose.

This work focuses on self-contained items assuming that the information they present is factual and comprehensive. In other applications such as healthcare, it is known that models struggle with data quality problems, such as missing, uncertain, or incorrect data \citep{arise2026report}. Realistically, managers also occasionally encounter the need to make decisions and plans based on low-quality data, which can be worth investigating in future work.

Only one frontier LLM was tested as AI judge in Section~\ref{ssec:automatic-metrics}, due to both time and space constraints. We do believe that the results presented are representative of the general behavior of AI judges, as we could perfectly replicate the behavior described in the literature at the first shot, but this calls for a more comprehensive analysis of the biases present in these LLM-based AI judges. For example: do less capable models tend to over-estimate the performance of more capable ones, or do they misestimate everything in general? Can more accurate assessments be derived from AI judges after accounting for their biases?

\section*{Ethical Considerations}

As this paper is focused on evaluation, we do not anticipate any malicious or unintended harmful effects of our work. The evaluation dataset was entirely composed of question items handcrafted by the recruited experts. All contributors were engaged under part-time, remote agreements across Europe with a fixed monthly stipend based on publicly available compensation data for early-career professional roles in their respective countries of residence. Contributors worked on a flexible schedule and were free to end their participation at any time. No personal information was involved in the work, except for the anonymized domain of expertise, which was handpicked by our team after a personal interview.

\section*{Acknowledgments}

We would like to thank Prof. Ehud Reiter for his invaluable feedback, as well as the participants in this study: Angad Singh, Anna Marie Augins, Anna Pokrovsky, Antonio Yutronic, Armando Nuno Leitão, Arun Chavan, Ashenafi Shiferaw, Benedic González, Bleriona Bahidi, Bruna Alves de Morais, Carolina Borlido, Christopher Gerbino, Deepam Sagar, Dheeraj Rajesh, Eldar Aliyev, Frederic Avis, Hripsime Hovsepyan, Ian D'Ambrosio, Kavya Velisetti, Kishor Murugan, Louise Andenmatten, Maria Jose Orejarena, Maria-Madalina Ifrim, Matteo Fabbri, Matteo Saffioti, Max Leal, Miguel Guerra, Nadin Khazendar, Natalia Quintero, Nguyen Tam An Vo, Nicolas Parra, Nini Makharadze, Ornella Romina Semino Arrieta, Pier Giacomo Agostinelli, Reyansh Seth, Robert Cheng, Salima Gouiaa, Sandhya Ravishankar, Sharad Sriram, Shalaka Lawrence, Sorina Gutu, Soumadeep Das, Sumant Duggal, Victor Musila, Vsevolods Mitrevics, and Zachary Miller.

\bibliography{custom}

\appendix

\section{Subsample with Expert Answers}
\label{app:subsample-expert-answers}

In Section \ref{sec:results} we present and analyze the results comparing the 6 models on 413 questions. For a subset of 33 items, we also asked the experts to write their own ideal answer in full, which was afterwards evaluated blindly together with the models by other domain experts. Table~\ref{tab:headline-33} contains the summary of the results from this subset. The model ranking from Section~\ref{sec:results} is clearly preserved, with some noise due to the small sample size (note that 1.5\% corresponds to a single instance). A fine-grained decomposition is not included, as the small sample size involves too much noise in any breakdown analysis.

\begin{table*}[t]  
    \centering
    \begin{tabular}{l|ccccc}
        \hline
        \textbf{Model} & \textbf{Solve Rate} & \textbf{Avg. Rank} & \textbf{Win (\%)} & \textbf{Rubric} & \textbf{Checkl. (\%)}  \\
        \hline
        Human Expert    & \textbf{92.4} & 1.76 & 74.24 & 4.33 & 94.9 \\
        Fable 5         & 39.4 & 2.80 & 9.1 & 4.04 & 54.8 \\
        GPT-5.5         & 33.3 & 3.68 & 6.1 & 3.80 & 51.0 \\
        Opus 4.8        & 21.2 & 4.14 & 3.0 & 3.62 & 46.6 \\
        Gemini 3.1 Pro  & 24.2 & 4.50 & 4.5 & 3.50 & 45.3 \\
        GLM-5.2         & 12.1 & 5.45 & 1.5 & 3.16 & 37.8 \\
        Mistral Large   & 13.6 & 5.67 & 1.5 & 3.05 & 36.9 \\
        \hline
    \end{tabular}
    \caption{Summary of results from the 33-item subset including blindly evaluated human responses}
    \label{tab:headline-33}
\end{table*}

\section{Monetary Cost}
\label{app:cost}

\begin{figure}[]
    \centering
    \includegraphics[width=\linewidth]{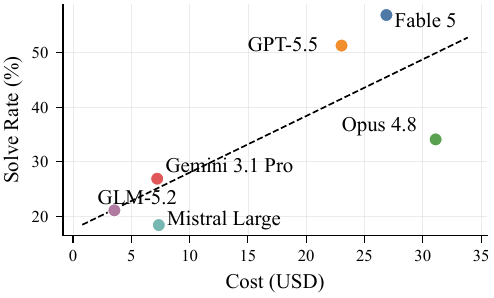}
    \caption{Cost of generating responses for all 413 questions against the ``Solve Rate'' aggregate metric}
    \label{fig:cost}
\end{figure}

The monetary cost of the models is not the main focus of this work, but it is a critical real-world factor that informs decisions. Figure \ref{fig:cost} plots the aggregate ``Solve Rate'' metric for every model against the cost of generating responses for the whole dataset, including a linear regression line. We can see a general tendency of more expensive models to perform better, but significant deviations exist especially at the higher end of the cost bracket.

\section{Response Length}
\label{app:length}

\begin{figure}[]
    \centering
    \includegraphics[width=\linewidth]{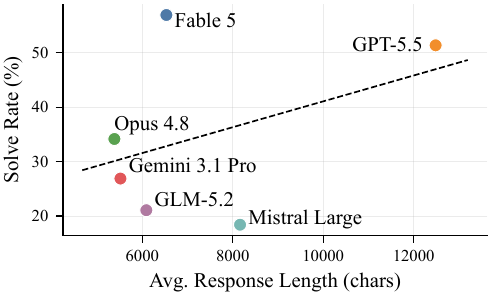}
    \caption{Length of responses against the ``Solve Rate'' aggregate metric}
    \label{fig:length}
\end{figure}

This appendix contains a secondary analysis about the average response length. As we can see in Figure~\ref{fig:length}, we observe no direct relationship between a model's response length and its ``Solve Rate''. At the item level, we see Pearson's $r$ values of 0.09 with the rubric and 0.11 with the checklist fulfillment rate. This small but positive value can be fully explained by the outlandishly extensive generations from GPT-5.5, which is one of the top-performers, and which incidentally can also explain its relatively poor assessments received in the Communication rubric attribute.

\section{Example of Graded Question Item}
\label{app:example-item}

I am CEO of a German grocery chain with 12 billion in annual revenue, an EBITDA margin of 6\%, the company is leveraged at 3.5 times their EBITDA and cannot exceed 4 times, and operates 2,000 stores across Germany, Austria, and the Czech Republic. The board has set a target EBITDA margin of 7\% while growing revenue 20\% over the next five years without becoming over-leveraged. 

You are considering expanding into Italy which has a highly fragmented grocery store environment. the top 5 players control 50\% of the environment and discount penetration is growing 6-7\% annually and average gross margin for discount stores is 2 percentage points higher in Germany because of more favorable conditions and less intense price competition. However labor costs are 20\% higher in Northern Italy than in Eastern Germany, Austria, and the Czech Republic. 

our strategy team has identified three options to reach the board's goals in the next 5 years:
1-open 300 stores in Northern Italy. this would cost 4 million euros for each store and requires building a new Italian distribution network at a cost of 250 million euros
2-a joint venture expansion into Italy with an existing mid-sized regional supermarket (owns 150 stores and 2.1 billion euros in revenue). the plan would convert 100 stores to discount stores and would cost 1 million euros for each store. the joint venture would pay profits 50\% once EBITDA exceeds 100 million
3-Germany first approach with small Italian test. focus 80\% of capex on automation, online ordering infrastructure, and format optimization in existing German stores. run a limited Italian pilot with 40 leased stores over 3 years 

write a one page memo for the supervisory board and recommend one of the options.

\begin{enumerate}
    \item States a clear recommendation (or a hybrid of two options) and explains why it rejects the other options with 1-2 reasons
    \item Quantifies Italy's 5 year contribution to revenue and EBITDA (for the recommended option) and clearly states how it moves the company towards the board's target of 20\% revenue growth and 7\% EBITDA margin
    \item Balances economic and social factors, specifically reputation and political risk from job reductions
    \item Identifies the top 3-5 risks (union backlash, tech underperformance, political pressure, cost overruns, etc.) and gives a concrete mitigation action for each option)
    \item Quantifies capex and EBITDA at the end of year 5 for chosen option (or hybrid option)
    \item Estimates final capex for chosen option at the end of each 5 years and is broken down into major components (Italian store openings, Italian distribution network, German automation, etc. for whichever option is recommended)
    \item Provides a comparison of labor costs impacts between entering Italy compared to a non-entry scenario, including change in total group labor cost
    \item Compares revenue, margin, capex, labor costs, EBITDA of chosen option to non-entry alternative hypothetical
\end{enumerate}

\paragraph{Model Response:} \textbf{Situation.} We generate €12B in revenue at a 6\% EBITDA margin (€720M), carrying €2.52B of debt (3.5x EBITDA). Our covenant ceiling of 4.0x leaves only ~€360M of immediate debt headroom. The Board's targets — €14.4B revenue and ~€1,008M EBITDA (7\% margin) by Year 5 — require us to grow and improve profitability simultaneously. The margin target is the harder of the two, and any expansion that dilutes margins or strains the balance sheet works against it.

\textbf{Assessment of Options.}\\ \textit{Option 1 — Build 300 stores in Northern Italy (€1.45B)}. This is financially unviable. The €1.45B requirement exceeds our debt headroom by roughly €1.1B and would push leverage toward 5.5x before new stores generate meaningful EBITDA (typically a 2–3 year ramp). The unit economics are also structurally inferior: Italian gross margins run ~2pp below Germany due to price competition, and Northern Italian labor costs are ~20\% above our core markets. Even at maturity, 300 Italian stores (~€1.8B revenue at our €6M/store average) would dilute the consolidated margin — moving us away from 7\%, not toward it. \textbf{Reject.}

\textit{Option 2 — Joint venture with Italian regional player (€100M).} Capital-light and attractive on the surface, but the economics are capped and the strategic control is not ours. The 50\% profit share above €100M EBITDA surrenders half the upside precisely where value is created; converted discount stores still face the same adverse gross margin and labor cost structure; and JV governance risk in a first-time market is high. The contribution to group EBITDA would be too small to move a €12B revenue base toward the 7\% target. \textbf{Reject as primary strategy} (though the partner remains a future acquisition or partnership candidate).

\textit{Option 3 — Germany-first with limited Italian pilot.} \textbf{Recommended.} Directing ~80\% of capex to automation, online ordering, and format optimization attacks the margin target directly, in the market where our gross margins are already 2pp more favorable. A 100bps margin improvement on our existing base is worth ~€120M+ of EBITDA — more than either Italian option delivers, at lower risk. The 40 leased-store pilot preserves optionality in a genuinely attractive market (fragmented, top-5 hold only 50\% share, discount growing 6–7\% annually) while capping downside: leases avoid capital lock-in, and three years of trading data will tell us whether Italian unit economics can clear our hurdle rates before we commit at scale.

\textbf{The honest risk: revenue.} Option 3 does not automatically deliver 20\% growth. We will close the gap through (a) like-for-like growth of 2–3\% annually driven by online and format upgrades (~€1.5–1.9B), (b) continued infill openings in our core three markets using residual capex and free cash flow, and (c) the Italian pilot as a Year 4–5 growth platform. If the pilot succeeds, our strengthened balance sheet and higher EBITDA base (raising 4.0x capacity to ~€4B) will fund a full Italian entry — potentially including acquisition of the Option 2 partner — from a position of strength.

\textbf{Recommendation.} Approve Option 3. It is the only path that hits the margin target, keeps leverage comfortably below covenant, and keeps Italy open as a de-risked second act rather than a leveraged bet on structurally weaker economics.

\textbf{Next steps:} Detailed capex allocation plan and pilot site selection for Board review within 60 days.

\paragraph{Human Evaluation:} represented in Table \ref{tab:grading-item}.

\begin{table}[h]
    \small
    \centering
    \begin{tabular}{ccccc|c}
        \hline
        \textbf{Dom.} & \textbf{Reas.} & \textbf{Act.} & \textbf{Comm.} & \textbf{Local.} & \textbf{Chkl.} \\
        \hline
        5 & 2 & 5 & 5 & 4 & $2/8$ \\
        3 & 1 & 2 & 2 & 1 & $1.5/8$ \\
        \hline
    \end{tabular}
    \caption{Pair of gradings for the assessed item}
    \label{tab:grading-item}
\end{table}

\filbreak
\section{Examples of Rejected Questions}
\label{app:validation-steps}

This Appendix contains examples of question items that were rejected at the separate stages of the validation pipeline for different reasons.

\subsection{Automatic QA}

\textbf{Rejected due to language}

I am the Business Development lead at a US-based start-up looking to break into the Spanish market. I sell a partnership software that essentially connects software providers with systems integrators to allow them to connect and understand which potential opportunities or customers or prospect they have in common. My solution then recommends which companies should partner together to increase their odds of closing deals (basically, allowing companies to partner with the right counterparts to win together).

As a start-up, I have limited budget and time to make a real impact and start getting revenues quickly. I have 100,000 euros in funding that I am prepared to spend on developing this market, and I need to know where to spend this money. I have 2 individuals working on my business development team, 1 marketing individuals, and 3 software engineers keeping the product up to date. All of my employees are currently based in the US, so I need to be strategic about where I send my BDs to hunt for prospects.

My solution is deployed on AWS as a SaaS offering. I am mainly targeting other small/medium sized companies (<100 employees) in the technology/systems integrator space.

Write a one-page summary that does two things: Size the Spanish market and develop a strategic plan for me to enter the Spanish market. Identify risks and potential gaps in my potential strategy. At the end of the document, include a single sentence that summarizes whether the strategy you described is internally-consistent.

\begin{enumerate}
    \item Identifies gap as the need for native Spanish speakers on team in order to conduct business in Spanish culture
    \item Calculates market size using a bottom-up, not top-down approachC
    \item Shows a clear ROI of how the 100,000 euros could generate potential revenues for the company
    \item Identifies risk as needing to deploy their software in AWS’ Spain region in order to meet ENS security requirements
    \item Identifies the Customer Acquisition Time at a minimum of 30 (20-30 new customers in 1 year is realistic)
\end{enumerate}

\textit{QA verdict: language}. Checklist rows 2, 5 are poorly written. Row 2 has a stray ``C'' appended (``bottom-up, not top-down approachC''), which appears to be a formatting error making the row slightly garbled. Row 5 is incoherent: ``Identifies the Customer Acquisition Time at a minimum of 30 (20-30 new customers in 1 year is realistic)'' conflates ``Customer Acquisition Time'' (a duration metric) with a count of new customers, making it unclear and self-contradictory — a grader could not reliably determine what is being tested. \\ \\
\textbf{Rejected due to checklist mismatch}

You are the CPO of TKO tacos in Spain. In order to launch a cost optimized menu for 2027 you are planning a digital only menu service that tracks what clients orders and matches it to their profile. You need to use the same menu for all of the venues across the country and the objective is to reduce the menu size to 5 dishes and 5 drinks. Your marketing team is concerned this will cause too much friction amongst customers coming into all venues. Forecast shows you can tolerate a 5\% decrease on demand and still maintain a 10\% increase in revenue due to the cost savings from optimizing the menu. How do you integrate this solution in order to keep a maximum 5\% decrease on demand?

\begin{enumerate}
    \item Recommends avoiding personal information
    \item Integrates the digital menu into a digital ordering solution in order to facilitate tracking
    \item Flags TKO as a chain, not needing to prioritize on customer experiences
    \item Identifies TKO is widely spread across Spain and reaching a consensus amongst all venues is not feasible
    \item Proposes integrating regional solutions rather than going national
\end{enumerate}

\textit{QA verdict: checklist mismatch}. Checklist rows 3, 4, 5 test things the prompt never asks for. Row 3 is problematic: it says TKO ``not needing to prioritize customer experience'' because it is a chain. This is a dubious conclusion contradicted by the prompt itself, which says the marketing team IS concerned about friction — the premise of the item is incorrect. Row 4 claims ``reaching a consensus amongst all venues is not feasible,'' but the prompt states the CPO’s objective IS to use the same menu for all venues, making this item contrary to the scenario. Row 5 proposes ``integrating regional solutions rather than going national,'' which directly contradicts the prompt’s explicit requirement of using the same menu nationally. Rows 3, 4, and 5 are either contradicted by the scenario or test conclusions the prompt never asks for and that conflict with the stated constraints. \\ \\
\textbf{Rejected due to hint leakage}

You are CTO of a Spanish industrial manufacturing firm (SA, Barcelona). Engineering is pushing to migrate our legacy on-premise assembly orchestration to a new “Serverless AI micro services” architecture hosted on AWS. The system will stream multi-model sensor data to the cloud to coordinate the timing sequences of our collaborative robotic arms (cobots). To guarantee low latency, the team is utilizing WebSockets and a Redis caching layer, achieving an impressive p99 latency of 45ms. Security has validated the architecture using zero-trust TLS 1.3. The CEO wants your sign-off tomorrow to deploy this serverless coordination to save on CapEx.

\begin{enumerate}
    \item Rejects Cloud orchestration: rejects the deployment, identifying that relying on non-deterministic cloud connections (WebSockets) physical cobot timing sequences introduces catastrophic physical safety risks
    \item Mandates Deterministic Edge: demands that all kinetic control loops remain on a local operational technology (OT) network utilizing real-time industrial protocols (EtherCAT), not standard web protocols
    \item Proposes Architectural Decoupling: proposes a hybrid pivot: authorizes the serverless cloud tier only for asynchronous anomaly detection, while isolating the real-time cobot actuation entirely to the edge
    \item Overrides the latency decoy: clarifies that a “p99 of 45ms” is actually unacceptable for industrial motion control, which requires sub – millisecond, highly deterministic jitter tolerances to prevent physical collisions
    \item Overrides the security decoy: notes that TLS 1.3 and zero-trust solve information security, but completely fail to address the operational resilience requirements of the EU machinery directive
\end{enumerate}

\textit{QA verdict: hint leakage}. Checklist items 1, 2, 4, 5 are revealed in the prompt. The prompt heavily telegraphs the expected answer: it explicitly mentions WebSockets (hinting at non-deterministic protocol rejection in item 1), p99 45ms latency (hinting at the latency decoy override in item 4), and TLS 1.3 / zero-trust (hinting at the security decoy override in item 5). The prompt frames these as ``impressive'' or ``validated'' achievements, but a model need only recognize these as the planted red herrings to produce checklist-matching answers without deep domain expertise. The on-premise-to-cloud migration framing also strongly hints at the edge/OT mandate in item 2.

\subsection{Difficulty Gate}

You are the CEO of a mid-sized US-based B2B SaaS company ("FieldOps Cloud") that provides:

\begin{itemize}
    \item Workforce scheduling
    \item Real-time dispatch optimisation
    \item AI-driven field-service automation
    \item Mobile technician workflow tools
\end{itemize}

Current scale: \$180M ARR. Strong US mid-market penetration (utilities, telecom field services, logistics). Recently expanded into UK with moderate success EBITDA positive but reinvesting heavily in growth. The company cannot hire more than 140 employees during the expansion period. The board requires EBITDA neutrality by the end of year 3. Expansion budget is capped at 55 million euros. No acquisitions are permitted. The CEO has committed to investors that only one European launch will occur before the next funding round. The board wants Germany. The CFO wants Portugal. Sales wants France. Legal refuse Italy. Engineering prefers Spain. Budget allows only one. The decision cannot be reversed for three years.

\begin{enumerate}
    \item Must avoid introducing facts not supported by the scenario or clearly identified as assumptions.
    \item Should recommend exactly one country and exactly one market entry model without proposing hybrid or sequential approaches
    \item Should demonstrate that the recommendation remains feasible within the 55 million euros investment cap and hiring limit
    \item Should apply a consistent decision standard when evaluating both the selected and the rejected options.
    \item Should maintain consistency between the recommend, supporting rationale, stated board constraints
\end{enumerate}

\textit{Average auto-fulfillment rate: 0.96}. See Table~\ref{tab:difficulty-reject} for each individual item's rate.

\begin{table}[h]
    \small
    \centering
    \begin{tabular}{l|ccccc}
        \hline
        \textbf{Item} & \textbf{1} & \textbf{2} & \textbf{3} & \textbf{4} & \textbf{5} \\
        \hline
        \textbf{Avg. Fulfillment} & 1 & 1 & 0.8 & 1 & 1 \\
        \hline
    \end{tabular}
    \caption{Average auto-fulfillment rate for every item of the checklist out of 5 runs}
    \label{tab:difficulty-reject}
\end{table}

\end{document}